\documentclass[final,5p,times,english,twocolumn]{elsarticle}

\usepackage{lineno}
\modulolinenumbers[5]

\journal{Neural Networks}

\usepackage{cite}

\usepackage{graphicx}

\usepackage{chngpage}
\usepackage{amsmath,amssymb,amsthm}
\usepackage{algorithm,algorithmic,amssymb,amstext,enumerate,amsfonts,url,multirow,array,stfloats,flushend,multirow}
\usepackage{caption}
\usepackage{booktabs}
\usepackage{threeparttable}

\usepackage[colorlinks,linkcolor=blue,citecolor=blue,urlcolor=blue]{hyperref}

\usepackage[T1]{fontenc}

\renewcommand\arraystretch{1.3}

\def\x{{\mathbf x}}
\def\w{{\mathbf w}}

\def\s{{\mathbf s}}

\def\h{{\mathbf h}}

\def\W{{\mathbf W}}

\def\H{{\mathbf H}}

\usepackage{amsmath}
\newtheorem{thm}{Theorem~}

\newtheorem{Cor}{Corollary}
\newtheorem{cor}[Cor]{Corollary}

\DeclareMathAlphabet{\mathcal}{OMS}{cmsy}{m}{n}

\newcolumntype{I}{!{\vrule width 3pt}}
\newlength\savedwidth

\newlength\savewidth
\newcommand\shline{\noalign{\global\savewidth\arrayrulewidth
                            \global\arrayrulewidth 1pt}%
                   \hline
                   \noalign{\global\arrayrulewidth\savewidth}}
\usepackage{amsmath,color}
\usepackage{algorithm}
\usepackage{algorithmic}

\usepackage{array}

\usepackage{bigstrut,bigdelim,multirow}

\biboptions{authoryear}

\begin{document}

\begin{frontmatter}

\title{Multilayer Bootstrap Networks}

\author{Xiao-Lei Zhang\corref{mycorrespondingauthor}}
\address{Center for Intelligent Acoustics and Immersive Communications, School of Marine Science and Technology, Northwestern Polytechnical University, China}
\cortext[mycorrespondingauthor]{Corresponding author}
\ead[url]{www.xiaolei-zhang.net}
\ead{xiaolei.zhang9@gmail.com, xiaolei.zhang@nwpu.edu.cn}


\begin{abstract}
Multilayer bootstrap network builds a gradually narrowed multilayer nonlinear network from bottom up for unsupervised nonlinear dimensionality reduction. Each layer of the network is a nonparametric density estimator. It consists of a group of k-centroids clusterings. Each clustering randomly selects data points with randomly selected features as its centroids, and learns a one-hot encoder by one-nearest-neighbor optimization.
Geometrically, the nonparametric density estimator at each layer projects the input data space to a uniformly-distributed discrete feature space, where the similarity of two data points in the discrete feature space is measured by the number of the nearest centroids they share in common. The multilayer network gradually reduces the nonlinear variations of data from bottom up by building a vast number of hierarchical trees implicitly on the original data space. Theoretically, the estimation error caused by the nonparametric density estimator is proportional to the correlation between the clusterings, both of which are reduced by the randomization steps.
\end{abstract}


\end{frontmatter}

\nolinenumbers

\section{Introduction}\label{sec:1}

 Principal component analysis (PCA) \citep{pearson1901lines} is a simple and widely used unsupervised dimensionality reduction method, which finds a coordinate system in the original Euclidean space that the linearly uncorrelated coordinate axes (called principal components) describe the most variances of data.
Because PCA is insufficient to capture highly-nonlinear data distributions, many dimensionality reduction methods are explored.

 {\color{black}{Dimensionality reduction has two core steps. The first step finds a suitable feature space where the density of data with the new feature representation can be well discovered, i.e. a density estimation problem. The second step discards the noise components or small variations of the data with the new feature representation, i.e. a principal component reduction problem in the new feature space.}}

 Dimensionality reduction methods are either linear \citep{niyogi2004locality} or nonlinear based on the connection between the data space and the feature space. This paper focuses on nonlinear methods, which can be categorized to three classes. The first class is kernel methods. It first projects data to a kernel-induced feature space, and then conducts PCA or its variants in the new space. Examples include kernel PCA \citep{scholkopf1998nonlinear}, Isomap \citep{tenenbaum2000global}, locally linear embedding (LLE) \citep{roweis2000nonlinear}, Laplacian eigenmaps \citep{shi2002normalized,ng2001spectral,belkin2003laplacian}, $t$-distributed stochastic neighbor embedding (t-SNE) \citep{van2008visualizing}, and their generalizations \citep{yan2007graph,nie2011spectral}. The second class is probabilistic models. It assumes that data are generated from an underlying probability function, and takes the posterior parameters as the feature representation. Examples include Gaussian mixture model and latent Dirichlet allocation \citep{blei2003latent}. The third class is autoassociative neural networks \citep{hinton2006reducing}. It
learns a piecewise-linear coordinate system explicitly by backpropagation, and uses the output of the bottleneck layer as the new representation.

However, the feature representations produced by the aforementioned methods are defined in continuous spaces. A fundamental weakness of using a continuous space is that it is hard to find a simple mathematical form that transforms the data space to an ideal continuous feature space, since a real-world data distribution may be non-uniform and irregular. To overcome this difficulty, a large number of machine learning methods have been proposed, such as distance metric learning \citep{xing2002distance} and kernel learning \citep{lanckriet2004learning} for kernel methods, and Dirichlet process prior for Baysian probabilistic models \citep{Teh05sharingclusters}, in which advanced optimization methods have to be applied. Recently, learning multiple layers of nonlinear transforms, named deep learning, is a trend \citep{hinton2006reducing}. A deep network contains more than one nonlinear layers. Each layer consists of a group of nonlinear computational units in parallel. Due to the hierarchical structure and distributed representation at each layer, the representation learning ability of a deep network is exponentially more powerful than that of a shallow network when given the same number of nonlinear units. However, the development of deep learning was mostly supervised, e.g.  \citep{schmidhuber2015deep,hinton2012deep,wang2017supervised,he2016deep,zhou2017deep,wang2017convolutional}. To our knowledge, deep learning for unsupervised dimensionality reduction seems far from explored \citep{hinton2006reducing}.

To overcome the aforementioned weakness in a simple way, we revisit the definition of frequentist probability for the density estimation subproblem of dimensionality reduction. \textit{Frequentist probability defines an event's probability as the limit of its relative frequency in a large number of trials} \citep{frequentistprobability}. In other words, the density of a local region of a probability distribution can be approximated by counting the events that fall into the local region. This paper focuses on exploring this idea. To generate the events, we resort to \textit{random resampling} in statistics  \citep{efron1979bootstrap,efron1993introduction}. To count the events, we resort to one-nearest-neighbor optimization and binarize the feature space to a discrete space.

 To further reduce the small variations and noise components of data, i.e. the second step of dimensionality reduction, we extend the density estimator to a gradually narrowed deep architecture, which essentially builds a vast number of hierarchical trees on the discrete feature space. The overall simple algorithm is named \textit{multilayer bootstrap networks} (MBN).

To our knowledge, although ensemble learning \citep{breiman2001random,freund1995desicion,friedman2000additive,dietterich2000ensemble,tao2006asymmetric}, which was triggered by random resampling, is a large family of machine learning, it is not very prevalent in unsupervised dimensionality reduction. Furthermore, we did not find methods that estimate the density of data in discrete spaces by random resampling, nor their extensions to deep learning.

This paper is organized as follows. In Section \ref{sec:2}, we describe MBN. In Section \ref{sec:geometry}, we give a geometric interpretation of MBN. In Section \ref{sec:3}, we justify MBN theoretically. In Section \ref{sec:4}, we study MBN empirically. In Section \ref{sec:motivate}, we introduce some related work. In Section \ref{sec:5}, we summarize our contributions. 

\section{Multilayer bootstrap networks}\label{sec:2}

\subsection{Network structure}

\begin{figure}[t]
\centering
\resizebox{6.22cm}{!}{\includegraphics*{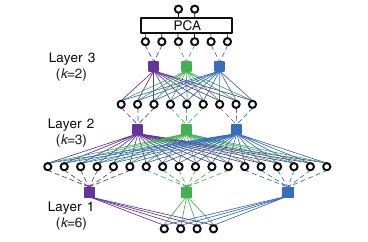}}
\caption{Network structure. The dimension of the input data for this demo network is 4. Each colored square represents a $k$-centroids clustering. Each layer contains 3 clusterings. Parameters $k$ at layers 1, 2, and 3 are set to 6, 3, and 2 respectively.
 The outputs of all clusterings in a layer are concatenated as the input of their upper layer.}
 \label{fig:1}
\end{figure}

 MBN contains multiple hidden layers and an output layer (Fig. \ref{fig:1}). Each hidden layer consists of a group of mutually independent $k$-centroids clusterings; each $k$-centroids clustering has $k$ output units, each of which indicates one cluster; the output units of all $k$-centroids clusterings are concatenated as the input of their upper layer. The output layer is PCA.

The network is gradually narrowed from bottom up, which is implemented by setting parameter $k$ as large as possible at the bottom layer and be smaller and smaller along with the increase of the number of layers until a predefined smallest $k$ is reached.

 \subsection{Training method}

  MBN is trained layer-by-layer from bottom up.

  For training each layer given a $d$-dimensional input data set $\mathcal{X} = \left\{\mathbf{x}_1,\ldots,\mathbf{x}_n \right\}$ either from the lower layer or from the original data space, we simply need to focus on training each $k$-centroids clustering, which consists of the following steps:

 \begin{itemize}
 \itemsep=0.0pt
   \item \textbf{Random sampling of features.} The first step randomly selects $\hat{d}$ dimensions of $\mathcal{X}$ ($\hat{d}\le d$) to form a subset of $\mathcal{X}$, denoted as $\hat{\mathcal{X}} = \left\{\hat{\mathbf{x}}_1,\ldots,\hat{\mathbf{x}}_n \right\}$.
    \item \textbf{Random sampling of data.} The second step randomly selects $k$ data points from $\hat{{\mathcal{X}}}$ as the $k$ centroids of the clustering, denoted as $\{\mathbf{w}_{1},\ldots,\mathbf{w}_{k} \}$.
   \item \textbf{One-nearest-neighbor learning.} The new representation of an input $\hat{\x}$ produced by the current clustering is an indicator vector $\mathbf{h}$ which indicates the nearest centroid of $\hat{\x}$. For example, if the second centroid is the nearest one to $\hat{\mathbf{x}}$, then $\mathbf{h} = [0,1,0,\ldots,0]^T$. The similarity metric between the centroids and $\hat{\mathbf{x}}$ at the bottom layer is customized, e.g. the squared Euclidean distance $ \arg\min_{i=1}^{k}\|\mathbf{w}_i-\hat{\mathbf{x}}\|^2$, and set to $\arg\max_{i=1}^{k}\mathbf{w}_i^T\hat{\mathbf{x}}$ at all other hidden layers.
 \end{itemize}

\subsection{Novelty and advantages}\label{subsec:novelty}
Two novel components of MBN distinguish it from other dimensionality reduction methods.

 The first component is that each layer is a nonparametric density estimator based on resampling, which has the following major merits:
 \begin{itemize}
   \item It estimates the density of data correctly without any predefined model assumptions. As a corollary, it is insensitive to outliers.
   \item The representation ability of a group of $k$-centroids clusterings is exponentially more powerful than that of a single $k$-centroids clustering.
   \item The estimation error introduced by binarizing the feature space can be controlled to a small value by simply increasing the number of the clusterings.
 \end{itemize}

 The second component is that MBN reduces the small variations and noise components of data by an unsupervised deep ensemble architecture, which has the following main merits:
   \begin{itemize}
   \item It reduces larger and larger local variations of data gradually from bottom up by building as many as $O(k_L2^V)$ hierarchical trees on the data space (instead of on data points) implicitly, where $L$ is the total number of layers, $k_L$ is parameter $k$ at the $L$-th layer, $V$ is the number of the clusterings at the layer, and function $O(\cdot)$ is the \textit{order} in mathematics.
   \item It does not inherit the problems of deep neural networks such as local minima, overfitting to small-scale data, and gradient vanishing, since it is trained simply by random resampling and stacking. As a result, it can be trained with as many hidden layers as needed and with both small-scale and large-scale data.
 \end{itemize}

 See Sections \ref{sec:geometry} and \ref{sec:3} for the analysis of the above properties.

 \subsection{Weaknesses}
 The main weakness of MBN is that the size of the network is large. Although MBN supports parallel computing, its prediction process is inefficient, compared to a neural network.

\begin{figure}[t]
\centering
\resizebox{8cm}{!}{\includegraphics*{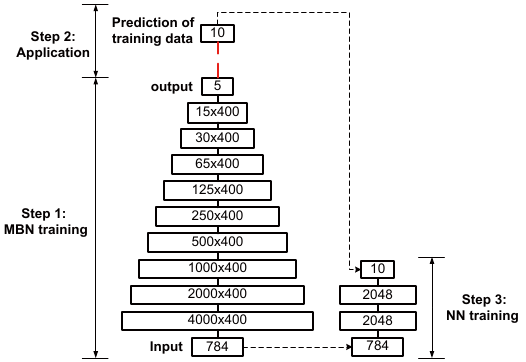}}
\caption{Principle of compressive MBN. The numbers are the dimensions of representations.}
 \label{fig:CMBN}
\end{figure}

 To overcome this weakness, we propose \textit{compressive MBN} (Fig. \ref{fig:CMBN}). Compressive MBN uses a neural network to learn a mapping function from the training data to the output of MBN at the training stage, and uses the neural network for prediction. More generally, if MBN is applied to some specific task, then we can use compressive MBN to learn a mapping function from the input data to the output of the task directly.

\section{Geometric interpretation}\label{sec:geometry}

In this section, we analyze the two components of MBN from the geometric point of view.

        \begin{figure*}[t]
 \centering
         \includegraphics[width=12cm]{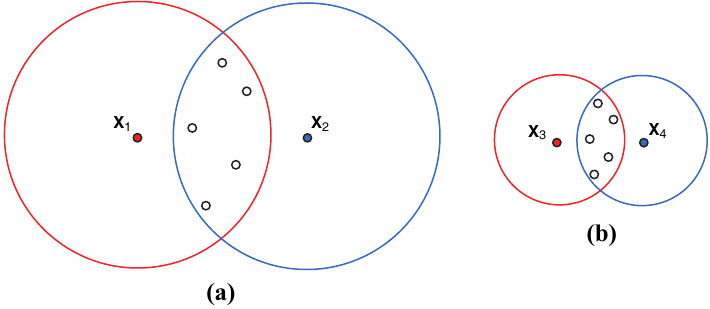}
 \caption{{{{Illustration of the similarity metric problem at the data space. \textbf{(a)} The similarity problem of two data points $\x_1$ (in red color) and $\x_2$ (in blue color) in a distribution $\mathcal{P}_1$. The local region of $\x_1$ (or $\x_2$) is the area in a colored circle that is centered at $\x_1$ (or $\x_2$). The small hollow points that lie in the cross area of the two local regions are the shared centroids by $\x_1$ and $\x_2$. \textbf{(b)} The similarity problem of two data points $\x_3$ and $\x_4$  in a distribution $\mathcal{P}_2$.}
 } }}
  \label{fig:binary}
 \end{figure*}

  \begin{figure*}[t]
 \centering
         \includegraphics[width=12cm]{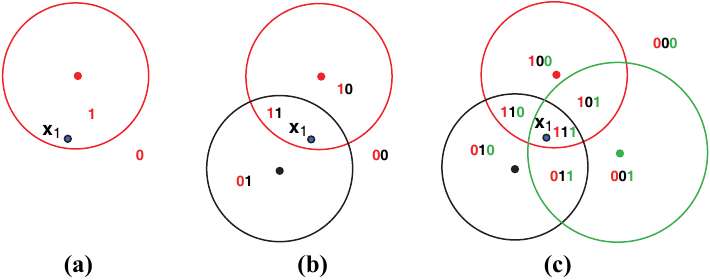}
 \caption{{{{Encoding the local region of a data point $\x_1$ (in blue color) by \textbf{(a)} one $k$-centroids clustering, \textbf{(b)} two $k$-centroids clusterings, and \textbf{(c)} three $k$-centroids clusterings. The centroids of the three $k$-centroids clusterings are the points colored in red, black, and green respectively. The local region of a centroid is the area in the circle around the centroid. The centroid itself and the edge of its local region are drawn in the same color. }
 } }}
  \label{fig:VQ}
 \end{figure*}

\subsection{Feature learning by a nonparametric density estimator based on resampling}\label{subsec:binarization}

As presented in Section \ref{sec:1}, a core problem of machine learning is to find a suitable similarity metric that maps the original data space to a uniformly-distributed feature space. Here we give an example on its importance. As shown in Fig. \ref{fig:binary}, the similarity between the data points that are far apart in a distribution with a large variance (e.g., $\x_1$ and $\x_2$ in a distribution $\mathcal{P}_1$) might be the same as the similarity between the data points that are close to each other in a distribution with a small variance (e.g., $\x_3$ and $\x_4$ in a distribution $\mathcal{P}_2$). If we use the Euclidean distance as the similarity metric, then the similarity between $\x_1$ and $\x_2$ is smaller than that between $\x_3$ and $\x_4$, which is not true.

The proposed method provides a simple solution to the above similarity metric learning problem. It outputs a new feature representation of $\x_1$ as follows. Each $k$-centroids clustering contributes a neighboring centroid of $\x_1$.
   The centroid partitions the local region of $\x_1$ to two disconnected parts, one containing $\x_1$ and the other not. The data point $\mathbf{x}_1$ owns a local region supported by the centroids of all clusterings that are closest to the data point.\footnote{It is important to keep parameter $k$ of the $k$-centroids clusterings in a layer the same. Otherwise, the density of the centroids between the clusterings are different, that is to say, if we regard the centroids as the coordinates axes of the local region, then the coordinate axes are built in different data spaces.}
   These centroids partition the local region to as many as $O(2^V)$ fractions. Each fraction is represented as a unique binary code at the output space of the estimator in a way illustrated in Fig. \ref{fig:VQ}.\footnote{A small fraction should not have to contain data points as shown in Fig. \ref{fig:VQ}.} The new representation of $\x_1$ in the output feature space is a binary code that represents the local fraction of the data space where $\x_1$ is located.

With the new representation, the similarity between two data points is calculated by counting the number of the nearest centroids they share in common---a method in frequentist methodology.
 In Fig. \ref{fig:binary}, (i) if the local region in $\mathcal{P}_1$ is partitioned in the same way as the local region in $\mathcal{P}_2$, and if the only difference between them is that the local region in $\mathcal{P}_1$ is an amplification of the local region in $\mathcal{P}_2$, then the surfaces of the two local regions are the same in the discrete feature space.
    (ii) $\x_1$ and $\x_2$ share 5 common nearest centroids, and $\x_3$ and $\x_4$ also share 5 common nearest centroids, so that the similarity between $\x_1$ and $\x_2$ in $\mathcal{P}_1$ equals to the similarity between $\x_3$ and $\x_4$ in $\mathcal{P}_2$.

    Note that because $V$ $k$-centroids clusterings are able to partition the data space to $O(k2^V)$ fractions at the maximum, its representation ability is exponentially more powerful than a single $k$-centroids clustering.

\subsection{Dimensionality reduction by a deep ensemble architecture}\label{subsec:deep_arch}
{\color{black}{
The nonparametric density estimator captures the variances of the input data, however, it is not responsible for reducing the small variances and noise components. To reduce the nonlinear variations, we build a gradually narrowed deep network. The network essentially reduces the nonlinear variations of data by building a vast number of hierarchical trees on the data space (instead of on data points) implicitly. We present its geometric principle as follows.

Suppose MBN has $L$ layers, parameters $k$ at layers $1$ to $L$ are $k_1,k_2,\ldots,k_L$ respectively, and $k_1>k_2>\ldots>k_L$. From Fig. \ref{fig:VQ}, we know that the $l$-th layer partitions the input data space to $O(k_l2^V)$ disconnected small fractions. Each fraction is encoded as a single point in the output feature space. The points in the output feature space are the nodes of the trees. Hence, the $l$-th layer has $O(k_l2^V)$ nodes. The bottom layer has $O(k_12^V)$ leaf nodes. The top layer has $O(k_L2^V)$ root nodes, which is the number of the trees that MBN builds.

For two adjacent layers, because $k_{l-1} < k_l$, it is easy to see that $O(k_{l-1}/k_l)$ neighboring child nodes at the $(l-1)$-th layer will be merged to a single father node at the $l$-th layer on average. To generalize the above property to the entire MBN, a single root node at the $L$-th layer is a merging of $O(k_1/k_L)$ leaf nodes at the bottom layer, which means that the nonlinear variations among the leaf nodes that are to be merged to the same root node will be reduced completely. Because $k_L\ll k_1$, we can conclude that MBN is highly invariant to the nonlinear variations of data.

One special case is that, if $k_L=1$, the output feature spaces of all $k$-centroids clusterings at the top layer are just a single point. However, in practice, we never set $k_L=1$; instead, we usually set the termination condition of MBN as $k_L\geq 1.5c$ where $c$ is the ground-truth number of classes. The termination condition makes each $k$-centroids clustering stronger than random guess \citep{schapire1990strength}.

Note that because $O(k2^V)\gg n$, the data points are distributed sparsely in the output feature space. Hence, MBN seldom merges data points. In other words, MBN learns data representations instead of doing agglomerative data clustering.}}

\section{Theoretical analysis}\label{sec:3}
   Although MBN estimates the density of data in the discrete feature space, its estimation error is small and controllable. Specifically, as shown in Fig. \ref{fig:VQ}c, $V$ $k$-centroids clusterings can partition a local region to $O(2^V)$ disconnected small fractions at the maximum. It is easy to imagine that, for any local region, when $V$ is increasing and the diversity between the centroids is still reserved, an ensemble of $k$-centroids clusterings approximates the true data distribution. Because the diversity is important, we have used two randomized steps to enlarge it.

In the following, we analyze the estimation error of the proposed method formally:
\begin{thm}\label{thm:0}
The estimation error of the building block of MBN $\mathcal{E}_{\scriptsize\mbox{ensemble}}$ and the estimation error of a single $k$-centroids clustering $\mathcal{E}_{\scriptsize\mbox{single}}$ have the following relationship:
 \setlength{\arraycolsep}{0.2em}
   \begin{eqnarray}
\mathcal{E}_{\scriptsize\mbox{ensemble}} = \left(\frac{1}{V}+\left(1-\frac{1}{V}\right)\rho\right)\mathcal{E}_{\scriptsize\mbox{single}}
 \end{eqnarray}
where $\rho$ is the pairwise positive correlation coefficient between the $k$-centroids clusterings, $0\leq \rho \leq 1$.
\end{thm}
\begin{proof}
We prove Theorem \ref{thm:0} by first transferring MBN to a supervised regression problem and then using the bias-variance decomposition of the mean squared error to get the bias and variance components of MBN. The detail is as follows.

Suppose the random samples for training the $k$-centroids clusterings are identically distributed but not necessarily independent, and
the \textit{pairwise positive correlation coefficient} between two random samples (i.e., $\{\w_{v_1,i}\}_{i=1}^k$ and $\{\w_{v_2,j}\}_{j=1}^k$, $\forall v_1,v_2=1,\ldots,V$ and $v_1\neq v_2$) is $\rho$, $0\leq\rho\leq 1$.

We focus on analyzing a given point $\x$, and assume that the \textit{true} local coordinate of $\x$ is $\s$ which is an invariant point around $\x$ and usually found when the density of the nearest centroids around $\x$ goes to infinity (i.e. $n\rightarrow \infty, V\rightarrow \infty$, $\{\w_v\}_{v=1}^V$ are identically and independently distributed, and parameter $k$ is unchanged). We also suppose that $\x$ is projected to $\hat{\s}$ when given a finite number of nearest centroids $\{\w_v\}_{v=1}^V$. The correlation coefficient between the centroids $\{\w_v\}_{v=1}^V$ is $\rho$. Note that $\s$ is used as an invariant reference point for studying $\hat{\s}$.

The effectiveness of MBN can be evaluated by the estimation error of the estimate $\hat{\s}$ to the truth $\s$. The \textit{estimation error} is defined by $E(\|\s-\hat{\s}\|^2)$ where $E(\cdot)$ is the expectation of a random variable.

 From the geometric interpretation, we know that each $\w_v$ owns a local space $S_v$, and moreover, both $\s$ and $\hat{\s}$ are in $S_v$. When parameter $k\rightarrow n$, $S_v$ is small enough to be locally linear. Under the above fact and an assumption that the features of $\w_v$, i.e. $w_{v,1},\ldots,w_{v,d},\ldots,w_{v,D}$, are uncorrelated, we are able to assume that $\w_v$ follows a multivariate normal distribution around $\s$: $\w_v\sim \mathcal{MN}(\s,\sigma^2\mathbf{I})$ where $\mathbf{I}$ is the identity matrix, $\sigma$ describes the variance of $\w_v$ and $\sigma^2\mathbf{I}$ is the covariance matrix.  We can further derive
$E(\|\s-\hat{\s}\|^2)=\sum_{d=1}^D E((s_d-\hat{s}_d)^2)$
 where we denote $\s=[s_1,\ldots, s_D]^T$ and $\hat{\s}=[\hat{s}_1,\ldots, \hat{s}_D]^T$.

 In the following, we focus on analyzing the estimation error at a single dimension $E((s_d-\hat{s}_d)^2)$ and will omit the index $d$ for clarity.
It is known that the mean squared error of a regression problem can be decomposed to the summation of a squared bias component and a variance component \citep{hastie2009unsupervised}:
 \setlength{\arraycolsep}{0.2em}
   \begin{eqnarray}
 E((s-\hat{s})^2)&=&\left( s - E\left(\hat{s}\right)\right)^2+ E\left(\left(\hat{s}-E\left(\hat{s}\right)\right)^2\right)\nonumber \\
&=& \text{Bias}^2\left(\hat{s}\right) + \text{Var}\left(\hat{s}\right)
\label{eq:12}
 \end{eqnarray}
Because $w_v$ follows a univariate normal distribution, it is easy to obtain:
   \setlength{\arraycolsep}{0.2em}
   \begin{eqnarray}
E(w_v)& =& s\label{eq:mean}\\
E(w_v^2) &=& \sigma^2 +s^2\label{eq:variance}\\
E(w_{v_1}w_{v_2})& = &\rho \sigma^2+s^2\label{eq:group_var}
 \end{eqnarray}
  For a single estimate $w_v$, we have $\hat{s}_{v}=w_v$. For a set of estimates $\{w_v\}_{v=1}^V$, we have $\hat{s}_{\Sigma} = \frac{1}{V}\sum_{v=1}^V w_v$. Based on Eqs. (\ref{eq:mean}) to (\ref{eq:group_var}), we can derive:
 \setlength{\arraycolsep}{0.2em}
    \begin{eqnarray}
\text{Bias}^2\left(\hat{s}_v\right) &= &0
\label{eq:theory5}
 \end{eqnarray}
    \begin{eqnarray}
\sigma^2_{\text{single}}=\text{Var}\left(\hat{s}_v\right) &= &\sigma^2
\label{eq:theory6}
 \end{eqnarray}
 and
     \setlength{\arraycolsep}{0.2em}
    \begin{eqnarray}
    \text{Bias}^2\left(\hat{s}_{\Sigma}\right) &= &0
\label{eq:theory7}
 \end{eqnarray}
    \begin{eqnarray}
\sigma^2_{\text{ensemble}}&=&\text{Var}\left(\hat{s}_{\Sigma}\right) =\frac{\sigma^2}{V}+\left(1-\frac{1}{V}\right)\rho\sigma^2.
\label{eq:theory8}
 \end{eqnarray}
Substituting Eqs. (\ref{eq:theory5}) and (\ref{eq:theory6}) to Eq. (\ref{eq:12}) gets:
 \setlength{\arraycolsep}{0.2em}
    \begin{eqnarray}
 \mathcal{E}_{\scriptsize\mbox{single}}&= &\sigma^2
\label{eq:err_v}
 \end{eqnarray}
 and substituting Eqs. (\ref{eq:theory7}) and (\ref{eq:theory8}) to Eq. (\ref{eq:12}) gets:
 \setlength{\arraycolsep}{0.2em}
    \begin{eqnarray}
 \mathcal{E}_{\scriptsize\mbox{ensemble}}&= &\frac{\sigma^2}{V}+\left(1-\frac{1}{V}\right)\rho\sigma^2.
\label{eq:err_ensemble}
 \end{eqnarray}
 Theorem \ref{thm:0} is proved.
\end{proof}

From Theorem \ref{thm:0}, we can get the following corollaries easily:
\begin{cor}\label{thm:1}
  When $\rho$ is reduced from $1$ to $0$, $\mathcal{E}_{{\scriptsize\mbox{ensemble}}}$ is reduced from  $\mathcal{E}_{{\scriptsize\mbox{single}}}$ to $\mathcal{E}_{{\scriptsize\mbox{single}}}/V$ accordingly.
\end{cor}
\begin{cor}\label{cor:2}
  When $V\rightarrow\infty$, $\mathcal{E}_{{\scriptsize\mbox{ensemble}}}$ reaches a lower bound  $\rho\mathcal{E}_{{\scriptsize\mbox{single}}}$.
\end{cor}
From Corollary \ref{thm:1}, we know that increasing parameter $V$ and reducing $\rho$ can help MBN reduce the estimation error.

From Corollary \ref{cor:2}, we know that it is important to reduce $\rho$. We have adopted two randomization steps to reduce $\rho$. However, although decreasing parameters $a$ and $k$ can help MBN reduce $\rho$, it will also cause $\mathcal{E}_{{\scriptsize\mbox{single}}}$ rise. In other words, reducing $\mathcal{E}_{{\scriptsize\mbox{single}}}$ and reducing $\rho$ is a pair of contradictory factors, which needs a balance through a proper parameter setting.

\section{Empirical evaluation}\label{sec:4}
In this section, we first introduce a typical parameter setting of MBN, then demonstrate the density estimation ability of MBN on synthetic data sets, and finally apply the low dimensional output of MBN to the tasks of visualization, clustering, and document retrieval.

 \subsection{Parameter setting}\label{subsec:typical}
 When a data set is small-scale, we use the linear-kernel-based kernel PCA \citep{scholkopf1998nonlinear,SVMKMToolbox} as the PCA toolbox of MBN.
When a data set is middle- or large-scale, we use the expectation-maximization PCA (EM-PCA) \citep{roweis1998algorithms}.\footnote{The word ``large-scale'' means that the data cannot be handled by traditional kernel methods on a common personal computer.}

\begin{table*}[htp]
\caption{\label{table:1}{Hyperparameters of MBN.}}
\renewcommand{\arraystretch}{1.5}
\centerline{\scalebox{0.9}{
\begin{threeparttable}
\begin{tabular}{m{0.15\textwidth} m{.78\textwidth}}
\hline
{Parameter} & {Description}\\
\shline
 $\delta$ & A parameter that controls the network structure by $k_{l+1}=\delta k_l$, $\forall l = 1,\ldots,L$\\
$a$ & Fraction of randomly selected dimensions (i.e., $\hat{d}$ ) over all dimensions (i.e., $d$) of input data.\\
$V$ & Number of $k$-centroids clusterings per layer.\\
 $k_1$ & A parameter that controls the time and storage complexities of the network. For small-scale problems, $k_1=0.5 n$. For large-scale problems, $k_1$ should be tuned smaller to fit MBN to the computing power.\\
\hline
\end{tabular}
\end{threeparttable}
}}
\end{table*}

MBN is insensitive to parameters $V$ and $a$ as if $V>100$ and $a\in[0.5,1]$. If not specified, we used $V=400$ and $a=0.5$ as the default values.

We denote parameter $k$ at layer $l$ as $k_l$. To control $k_l$, we introduce a parameter $\delta$ defined as $k_{l+1}=\delta k_l$, $\delta\in(0,1)$. MBN is relatively sensitive to parameter $\delta$: if data are highly-nonlinear, then set $\delta$ to a large value, otherwise, set $\delta$ to a small value; if the nonlinearity of data is unknown, set $\delta = 0.5$ which is our default.

Finally, if $k_{L+1}<1.5{c}$, then we stop MBN training and use the output of the $L$-th nonlinear layer for PCA. This terminating condition guarantees the validness of data resampling which requires each random sample of data to be stronger than random guess \citep{schapire1990strength}. The hyperparameters of MBN are summarized in Table \ref{table:1}.

{In the following experiments, we adopted the above default parameter setting of MBN, unless otherwise specified.}

\subsection{Density estimation}
 \subsubsection{Density estimation for nonlinear data distributions}

Four synthetic data sets with non-uniform densities and nonlinear variations are used for evaluation. They are Gaussian data, Jain data \citep{Jain2005data}, pathbased data \citep{Pathbased2008data}, and compound data \citep{Compound1971graph}, respectively. Parameter $a$ was set to 1.

\begin{figure*}[t]
 \centering
         \includegraphics[width=18cm]{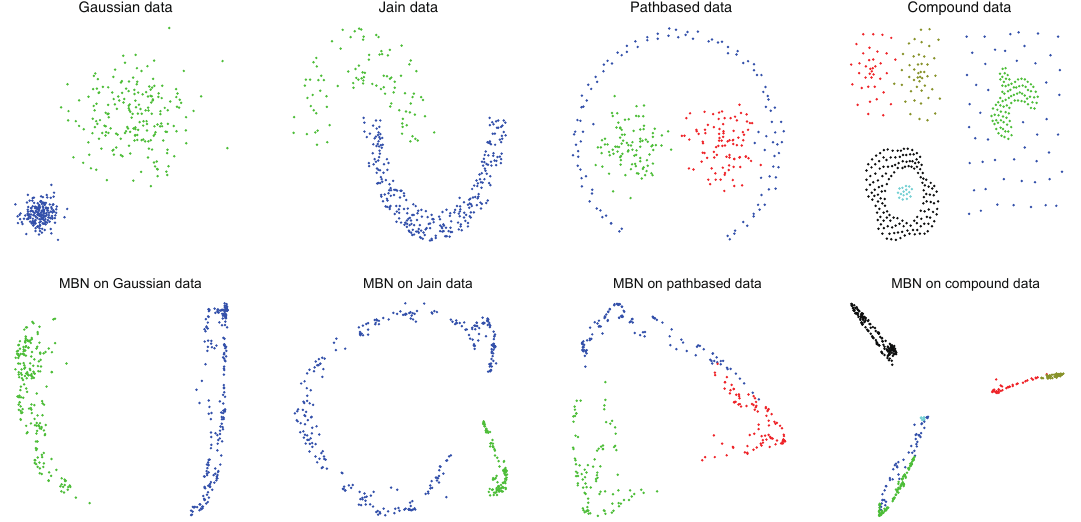}
 \caption{{{{Density estimation by MBN on synthetic data sets}. The data points in different classes are drawn in different colors.}
 } }
  \label{fig:toy_visualization}
 \end{figure*}

The visualization result in Fig. \ref{fig:toy_visualization} shows that the synthetic data in the new feature spaces produced by MBN not only are distributed uniformly but also do not contain many nonlinear variations.

 \subsubsection{Density estimation in the presence of outliers}
\begin{figure}[t]
 \centering
         \includegraphics[width=7.5cm]{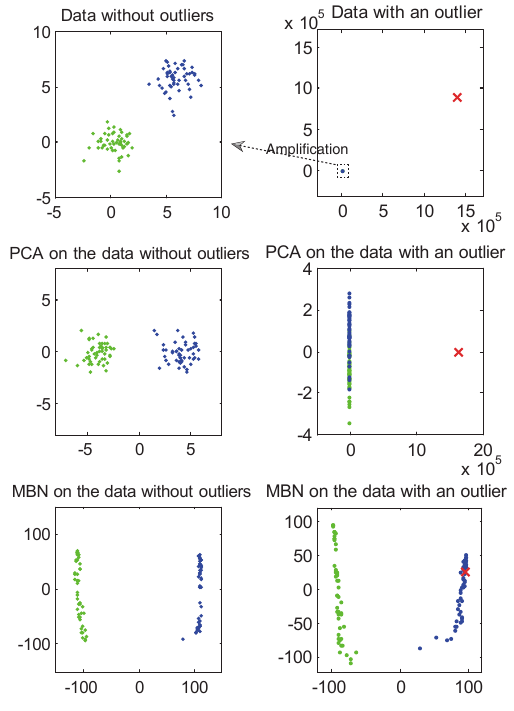}
 \caption{{{{Density estimation by MBN on a Gaussian data with the presence of an outlier}. The red cross denotes the outlier. Because the outlier is far from the Gaussian data, the resolution of the sub-figure in the top-right corner is not high enough to differentiate the two classes of the Gaussian data. }
 } }
  \label{fig:outliers}
 \end{figure}

 A data set of two-class Gaussian data with a randomly generated outlier is used for evaluation. The results of PCA and MBN are show in Fig. \ref{fig:outliers}. From the figure, we observe that MBN is robust to the presence of the outlier.

\subsection{Data visualization}\label{subsec:xxMNIST}

\subsubsection{Visualizing AML-ALL biomedical data}\label{subsec:small_AMLALL}

The acute myeloid leukemia and acute lymphoblastic leukemia (AML-ALL) biomedical data set \citep{golub1999molecular} is a two-class problem that consists of 38 training examples (27 ALL, 11 AML) and 34 test examples (20 ALL, 14 AML). Each example has 7,129 dimensions produced from 6,817 human genes.

\begin{figure*}[t]
 \centering
         \includegraphics[width=18cm]{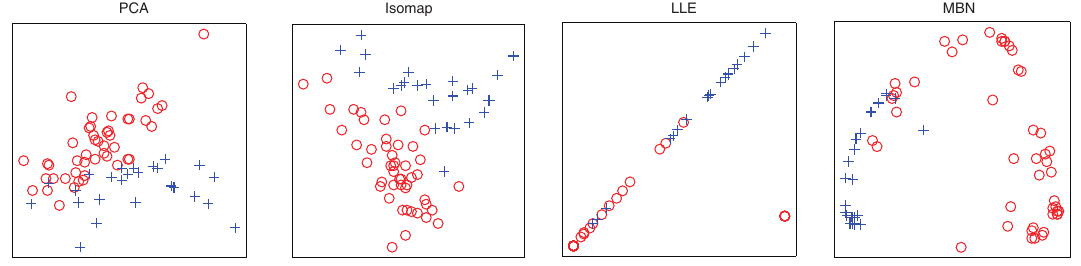}
 \caption{{{{Visualizations of AML-ALL produced by MBN and 3 comparison methods.}
 } }}
  \label{fig:amlall_visualization}
 \end{figure*}

We compared MBN with PCA and 2 nonlinear dimensionality reduction methods which are Isomap \citep{tenenbaum2000global} and LLE \citep{roweis2000nonlinear}. We tuned the hyperparameters of Isomap and LLE for their best performance.\footnote{It is difficult and sometimes unable to tune the hyperparameters in practice.}
The visualization result is shown in Fig. \ref{fig:amlall_visualization}.

\subsubsection{Visualizing MNIST digits}\label{subsec:small_MNIST}


     \begin{figure*}
 \centering
         \includegraphics[width=18cm]{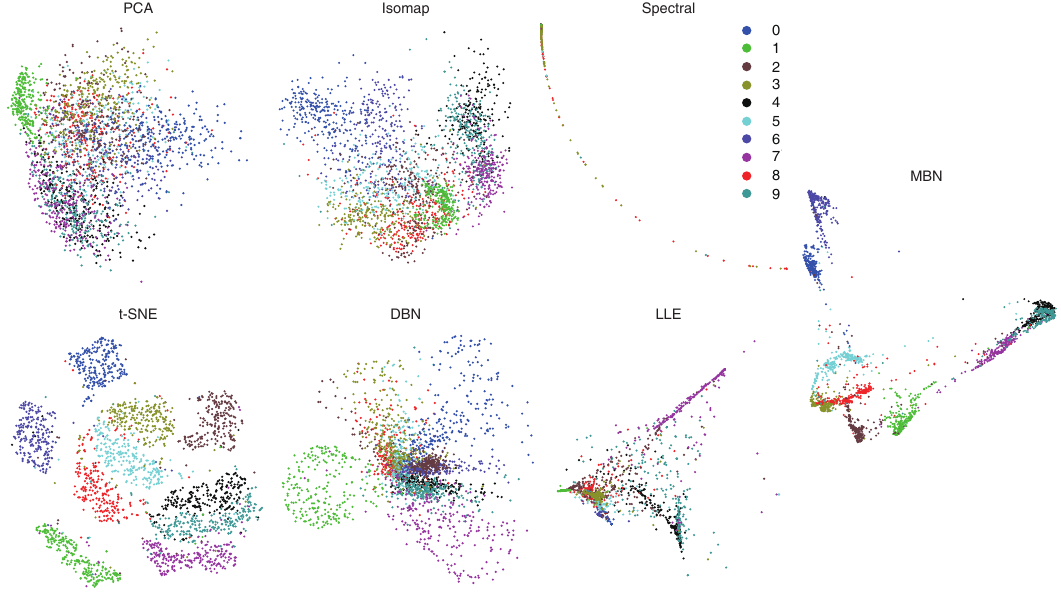}
 \caption{{{{Visualizations of a subset (5,000 images) of MNIST produced by MBN and 6 comparison methods. For clarity, only 250 images per digit are drawn.}
 } }}
  \label{fig:3}
 \end{figure*}

   \begin{figure}[t]
\centering
{\includegraphics[width=7cm]{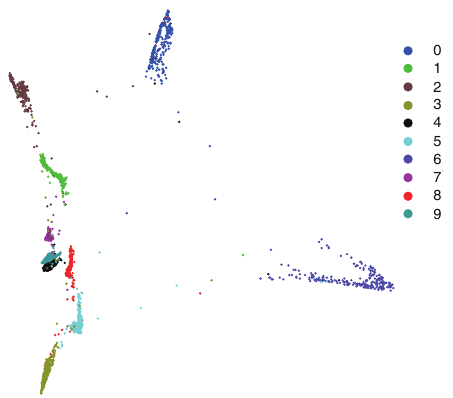}}
\caption{Visualization of the 10,000 test images of MNIST produced by the MBN at layer 8. For clarity, only 250 images per digit are drawn. See \ref{app:c} for the visualizations produced by other layers.}
 \label{fig:MNIST_visualization}
\end{figure}

 The data set of the MNIST digits \citep{MNIST} contains 10 handwritten integer digits ranging from 0 to 9. It consists of 60,000 training images and 10,000 test images. Each image has 784 dimensions.

    We compared MBN with PCA, Isomap \citep{tenenbaum2000global}, LLE \citep{roweis2000nonlinear}, Spectral \citep{ng2001spectral}, deep belief networks \citep{hinton2006reducing} (DBN), and t-SNE \citep{van2008visualizing}. We tuned the hyperparameters of the 5 nonlinear comparison methods for their best performance.

    Because the comparison nonlinear methods are too costly to run with the full MNIST data set except DBN, we randomly sampled 5,000 images with 500 images per digit for evaluation. The visualization result in Fig. \ref{fig:3} shows that the low-dimensional feature produced by MBN has a small within-class variance and a large between-class distance.

To demonstrate the scalability of MBN on larger data sets and its generalization ability on unseen data, we further trained MBN on all 60,000 training images, and evaluated its effectiveness on the 10,000 test images, where $k_1$ was set to $8000$ to reduce the training cost. The visualization result in Fig. \ref{fig:MNIST_visualization} shows that MBN on the full MNIST provides a clearer visualization than that on the small subset of MNIST.

\subsection{Clustering}\label{subsec:16data}

 Ten benchmark data sets were used for evaluation. They cover topics in speech processing, chemistry, biomedicine, image processing, and text processing. The details of the data sets are given in Table \ref{table:data_set_info}. The data set ``20-Newsgroups'', which originally has 20,000 documents, was post-processed to a corpus of 18,846 documents, each belonging to a single topic.

We compared MBN with PCA and Spectral \citep{ng2001spectral}. We applied $k$-means clustering to the low-dimensional outputs of all comparison methods as well as the original high-dimensional features. To prevent the local minima problem of $k$-means clustering, we ran $k$-means clustering 50 times and picked the clustering result that corresponded to the optimal objective value of the $k$-means clustering among the 50 candidate objective values as the final result.  We ran each comparison method 10 times and reported the average performance.

\begin{table*}[t]
\caption{\label{table:data_set_info}{Description of data sets.}}
\renewcommand{\arraystretch}{1.5}
\centerline{\scalebox{0.9}{
\begin{threeparttable}
\begin{tabular}{llllll}
\hline
ID &{Name} & {\# data points} & \# dimensions & \# classes & Attribute\\
 \shline
1& Isolet1 & 1560 & 617 & 26 & Speech data\\
2& Wine & 178 & 13 & 3 & Chemical data\\
3& New-Thyroid & 215 & 5 & 3 & Biomedical data\\
4& Dermathology & 366 & 34 & 6 & Biomedical data\\
5& COIL100 & 7200   & 1024  & 100 & Images\\
6& MNIST(small) & 5000  & 768 & 10 & Images (handwritten digits)\\
7& MNIST(full) & 70000  & 768 & 10 & Images (handwritten digits)\\
8& UMIST & 575 & 1024 & 20 & Images (faces)\\
9& ORL & 400  & 1024 &40 & Images (faces)\\
10& 20-Newsgroups & 18846   & 26214  &20 & Text corpus\\
\hline
\end{tabular}
\end{threeparttable}
}}
\end{table*}

   \begin{table*}[t]
\caption{\label{table:clustering results}{NMI on 10 data sets. The numbers after $\pm$ are the standard deviations.}}
\renewcommand{\arraystretch}{1.5}
\centerline{\scalebox{0.9}{
\begin{tabular}{llllll|ll}
\hline
& & {Raw feature} & PCA & Spectral$^{no\_tuning}$& MBN$^{no\_tuning}$ & Spectral{$^{optimal}$}& MBN{$^{optimal}$}\\
 \shline
1& Isolet1 & \textbf{77.21\%$\pm$0.92\%}& 56.74\%$\pm$0.75\%&  75.51\%$\pm$0.58\% &\textbf{77.89\%$\pm$0.81\%} & \textbf{79.66\%$\pm$0.58\%} & 77.97\%$\pm$0.64\%\\
2& Wine&  42.88\%$\pm$0.00\%& 40.92\%$\pm$0.00\%&    41.58\%$\pm$0.00\%    &\textbf{55.49\%$\pm$4.07\%}& 43.02\%$\pm$0.00\% & \textbf{76.96\%$\pm$3.24\%}\\
3& New-Thyroid&  49.46\%$\pm$0.00\%& 49.46\%$\pm$ 0.00\%&  39.63\%$\pm$0.00\%     &\textbf{68.80\%$\pm$5.03\%}& 43.20\%$\pm$0.00\% & \textbf{75.78\%$\pm$2.75\%}\\
4& Dermathology & 9.11\%$\pm$0.11\%& 59.50\%$\pm$0.10\%&   51.04\%$\pm$0.15\%     &\textbf{82.40\%$\pm$2.24\%}& 51.04\%$\pm$0.15\% & \textbf{92.79\%$\pm$1.09\%}\\
5& COIL100 & 76.98\%$\pm$0.27\%& 69.64\%$\pm$0.45\%&   \textbf{89.28\%$\pm$0.59\%}       &{81.66\%$\pm$0.59\%}& 89.28\%$\pm$0.59\% & \textbf{90.37\%$\pm$0.65\%}\\
6& MNIST(small) & 49.69\%$\pm$0.14\%& 27.86\%$\pm$0.08\%&    62.06\%$\pm$0.04\%     &\textbf{77.12\%$\pm$0.35\%}& 62.06\%$\pm$0.04\% & \textbf{78.50\%$\pm$0.84\%}\\
7& MNIST(full) &50.32\%$\pm$0.12\%& 28.05\%$\pm$0.07\% &     Timeout    & \textbf{91.36\%$\pm$0.07\%}& Timeout & Timeout\\
8& UMIST & 65.36\%$\pm$1.21\%& 66.25\%$\pm$1.10\%&    \textbf{81.83\%$\pm$0.58\%}     &{74.48\%$\pm$1.91\%}& \textbf{84.66\%$\pm$2.38\%} &\textbf{84.84\%$\pm$1.98\%}\\
9& ORL & 75.55\%$\pm$1.36\%& 75.81\%$\pm$1.17\%&    \textbf{80.21\%$\pm$1.13\%}        &\textbf{79.22\%$\pm$0.84\%}& \textbf{83.62\%$\pm$1.13\%} &81.73\%$\pm$1.19\%\\
10& 20-Newsgroups& Timeout & 22.49\%$\pm$1.41\%&   27.03\%$\pm$0.05\%      &\textbf{41.61\%$\pm$0.05\%}& 27.03\%$\pm$0.05\% &\textbf{42.23\%$\pm$0.46\%}\\
\hline
\end{tabular}
}}
\end{table*}

   \begin{table*}[t]
\caption{\label{table:clustering results_ACC}{Clustering accuracy on 10 data sets.}}
\renewcommand{\arraystretch}{1.5}
\centerline{\scalebox{0.9}{
\begin{threeparttable}
\begin{tabular}{llllll|ll}
\hline
& & {Raw feature} & PCA & Spectral$^{no\_tuning}$& MBN$^{no\_tuning}$ & Spectral{$^{optimal}$}& MBN{$^{optimal}$}\\
 \shline
1& Isolet1 & \textbf{61.47\%$\pm$1.93\%}& 38.62\%$\pm$0.99\%& 49.63\%$\pm$1.88    &\textbf{61.13\%$\pm$2.52\%} & \textbf{72.29\%$\pm$1.92\%} & 62.50\%$\pm$1.14\% \\
2& Wine& 70.22\%$\pm$0.00\%& 78.09\%$\pm$0.00\%&  61.24\%$\pm$0.00     &\textbf{81.91\%$\pm$2.61\%} & 70.79\%$\pm$0.00\% & \textbf{93.20\%$\pm$1.77\%} \\
3& New-Thyroid& 86.05\%$\pm$0.00\%& 86.05\%$\pm$0.00\%& 79.49\%$\pm$0.94       &\textbf{93.02\%$\pm$1.60\%} & 82.79\%$\pm$0.00\% & \textbf{94.51\%$\pm$0.93\%}\\
4& Dermathology &26.17\%$\pm$0.28\%& 61.67\%$\pm$0.26\%&  50.38\%$\pm$0.35     & \textbf{82.81\%$\pm$7.67\%} & 50.38\%$\pm$0.35\% & \textbf{96.07\%$\pm$0.79\%}\\
5& COIL100 &49.75\%$\pm$1.31\%& 43.42\%$\pm$1.21\%&  \textbf{61.83\%$\pm$2.27}     & {57.38\%$\pm$1.85\%} &  61.83\%$\pm$2.27\% & \textbf{68.29\%$\pm$1.58\%}\\
6& MNIST(small) & 52.64\%$\pm$0.14\%& 34.49\%$\pm$0.10\%&  53.30\%$\pm$0.01      &\textbf{82.36\%$\pm$0.46\%} & 56.55\%$\pm$0.20\% & \textbf{87.06\%$\pm$1.75\%}\\
7& MNIST(full) & 53.48\%$\pm$0.11\% & 35.14\%$\pm$0.11\%&   Timeout    & \textbf{96.64\%$\pm$0.04\%} &  Timeout & Timeout\\
8& UMIST & 43.20\%$\pm$1.66\%& 43.44\%$\pm$1.92\%&  \textbf{70.99\%$\pm$2.19}      &{56.96\%$\pm$3.73\%} & 70.99\%$\pm$2.19\%  & \textbf{73.22\%$\pm$4.02\%}\\
9& ORL &54.37\%$\pm$2.41\%&  54.55\%$\pm$2.81\%& 57.33\%$\pm$2.37       &\textbf{59.68\%$\pm$1.58\%} & \textbf{68.85\%$\pm$2.48\%} & 64.25\%$\pm$3.50\%\\
10& 20-Newsgroups&Timeout &22.61\%$\pm$1.29\% &  28.52\%$\pm$0.03      &\textbf{46.57\%$\pm$1.10\%} & 28.52\%$\pm$0.03\%  & \textbf{47.69\%$\pm$0.92\%}\\
\hline
\end{tabular}
\end{threeparttable}
}}
\end{table*}

The parameter settings on the data sets with IDs from 1 to 9 are as follows. For PCA, we preserved the top 98\% largest eigenvalues and their corresponding eigenvectors. For Spectral, we set the output dimension to the ground-truth number of classes and adopted the RBF kernel. We reported the results with a fixed kernel width $2^{-4}A$ which behaves averagely the best over all data sets, as well as the best result by searching the kernel width from $\{2^{-4}A,2^{-3}A,\ldots,2^4A\}$ on each data set, where $A$ is the average pairwise Euclidean distance between data. The two parameter selection methods of Spectral are denoted as Spectral$^{no\_tuning}$ and Spectral$^{optimal}$ respectively. For MBN, we set the output dimension to the ground-truth number of classes. We reported the results with $\delta=0.5$, as well as the best results by searching $\delta$ from $\{0.1,0.2,\ldots, 0.9\}$ on each data set. The two parameter selection methods of MBN are denoted as MBN$^{no\_tuning}$ and MBN$^{optimal}$ respectively.

The parameter settings on the 20-Newsgroups are as follows. For PCA, we set the output dimension to 100. For all comparison methods, we used the cosine similarity as the similarity metric.

We evaluated the clustering result in terms of normalized mutual information (NMI) and clustering accuracy. The clustering results in Tables \ref{table:clustering results} and \ref{table:clustering results_ACC} show that (i) MBN$^{no\_tuning}$ achieves better performance than Spectral$^{no\_tuning}$ and PCA, and (ii) MBN$^{optimal}$ achieves better performance than Spectral$^{optimal}$.

{\color{black}{Besides the data sets in Table \ref{table:data_set_info}, we have also conducted experiments on the following data sets: Lung-Cancer biomedical data, COIL20 images, USPS images, Extended-YaleB images, Reuters-21578 text corpus, and TDT2 text corpus. The experimental conclusions are consistent with the results in Tables \ref{table:clustering results} and \ref{table:clustering results_ACC}.}}

\begin{figure*}[t]
 \centering
         \includegraphics[width=12cm]{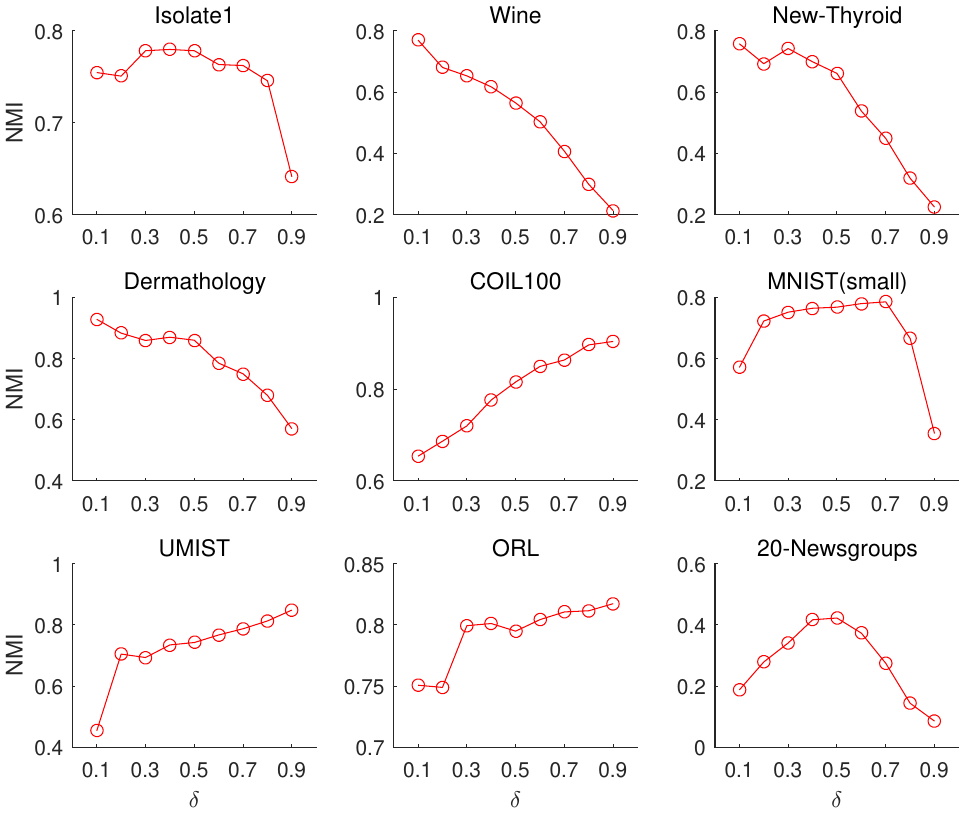}
 \caption{{{Effect of decay factor $\delta$ on 9 benchmark data sets.
 } }}
  \label{fig:16datasets}
 \end{figure*}

\subsubsection{Effect of hyperparameter $\delta$}\label{subsec:analysis_rho}

We showed the effect of parameter $\delta$ on each data set in Fig. \ref{fig:16datasets}. From the figure, we do not observe a stable interval of $\delta$ where MBN is supposed to achieve the optimal performance across the data sets; if the data are highly variant, then setting $\delta$ to a large value yields good performance, and vise versa. Generally, if the nonlinearity of data is unknown, then setting $\delta=0.5$ is safe.

  \begin{figure}[t]
 \centering
         \includegraphics[width=8.7cm]{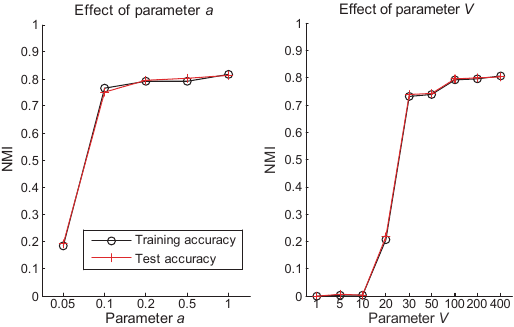}
         \caption{{Effect of parameters $a$ and $V$ on the subset of MNIST.}}
         \label{fig:param_V}
 \end{figure}

\subsubsection{Effects of hyperparameters $a$ and $V$}

Theorem \ref{thm:1} has guaranteed that the estimation error can be reduced by enlarging parameter $V$, and may also be reduced by decreasing parameter $a$. In this subsection, we focus on investigating the stable working intervals of $V$ and $a$.

Because the experimental phenomena on all data sets are similar, we reported the phenomena on the small subset of MNIST as a representative in Fig. \ref{fig:param_V}. From the figure, we know that (i) we should prevent setting $a$ to a very small value. Empirically, we set $a=0.5$. (ii) We should set $V\geq 100$. Empirically, we set $V= 400$.

%

\subsection{Document retrieval}\label{subsec:RCV}

      \begin{figure*}[htp]
 \centering
         \includegraphics[width=14cm]{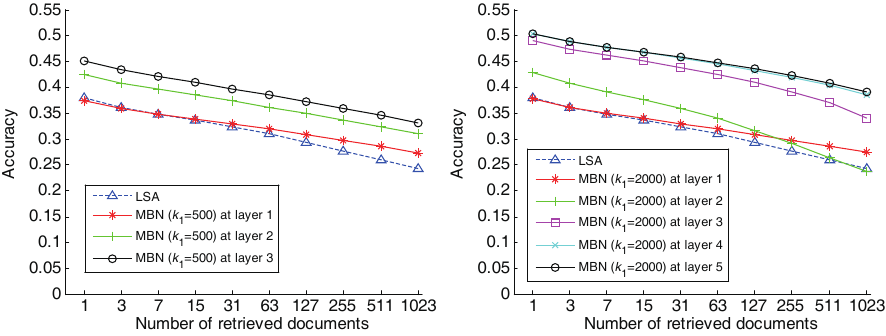}
 \caption{{{{Average accuracy curves of retrieved documents on the test set (82 topics) of the Reuters newswire stories.}
 } }}
  \label{fig:rcv_acc}
 \end{figure*}

We applied MBN to document retrieval and compared it with latent semantic analysis (LSA) \citep{deerwester1990indexing}, a document retrieval method based on PCA, on a larger data set---Reuters newswire stories \citep{lewis2004rcv1} which consist of 804,414 documents.
The data set of the Reuters newswire stories are divided into 103 topics which are grouped into a tree structure. We only preserved the 82 leaf topics. As a result, there were 107,132 unlabeled documents.
 We preprocessed each document as a vector of 2,000 commonest word stems by the \textit{rainbow} software \citep{rainbow} where each entry of a vector was set to the word count.

 We randomly selected half of the data set for training and the other half for test. We recorded the average accuracy over all 402,207 queries in the test set at the document retrieval setting, where a query and its retrieved documents were different documents in the test set.
 If an unlabeled document was retrieved, it was considered as a mistake. If an unlabeled document was used as a query, no relevant documents would be retrieved, which means the precisions of the unlabeled query at all levels were zero.

Because the data set is relatively large, we did not adopt the default setting of MBN where $k_1=201103$. Instead, we set $k_1$ manually to $500$ and $2000$ respectively, $V=200$, and kept other parameters unchanged, i.e. $a=0.5$ and $\delta = 0.5$.

 Experimental results in Fig. \ref{fig:rcv_acc} show that the MBN with the small network reaches an accuracy curve of over 8\% higher than LSA; the MBN with the large network reaches an accuracy curve of over 13\% higher than LSA. The results indicate that enlarging the network size of MBN improves its generalization ability.

\subsection{Empirical study on compressive MBN}\label{subsec:compressiveMBN}
This subsection studies the effects of the compressive MBN on accelerating the prediction process of MBN.

\begin{figure}[t]
 \centering
       \includegraphics[width=8.7cm]{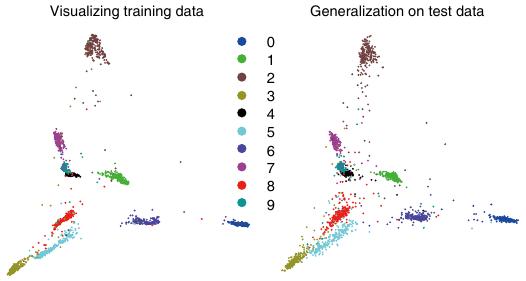}
 \caption{{Visualizations of the training data (left) and test data (right) of MNIST produced by the compressive MBN.}}
 \label{fig:visualization_MNIST_cMBN_without}
 \end{figure}

We first studied the generalization ability of the compressive MBN on the 10,000 test images of MNIST, where the models of MBN and the compressive MBN are trained with the 60,000 training images of MNIST. The neural network in the compressive MBN contains 2 hidden layers with 2048 rectified linear units per layer. It projects the data space to the 2-dimensional feature space produced by MBN.
The visualization results in Fig. \ref{fig:visualization_MNIST_cMBN_without} show that the compressive MBN produces an identical $2$-dimensional feature with MBN on the training set and generalizes well on the test set.
The prediction time of MBN and the compressive MBN is 4857.45 and 1.10 seconds, respectively.

\begin{figure}[t]
 \centering
         \includegraphics[width=6.3cm]{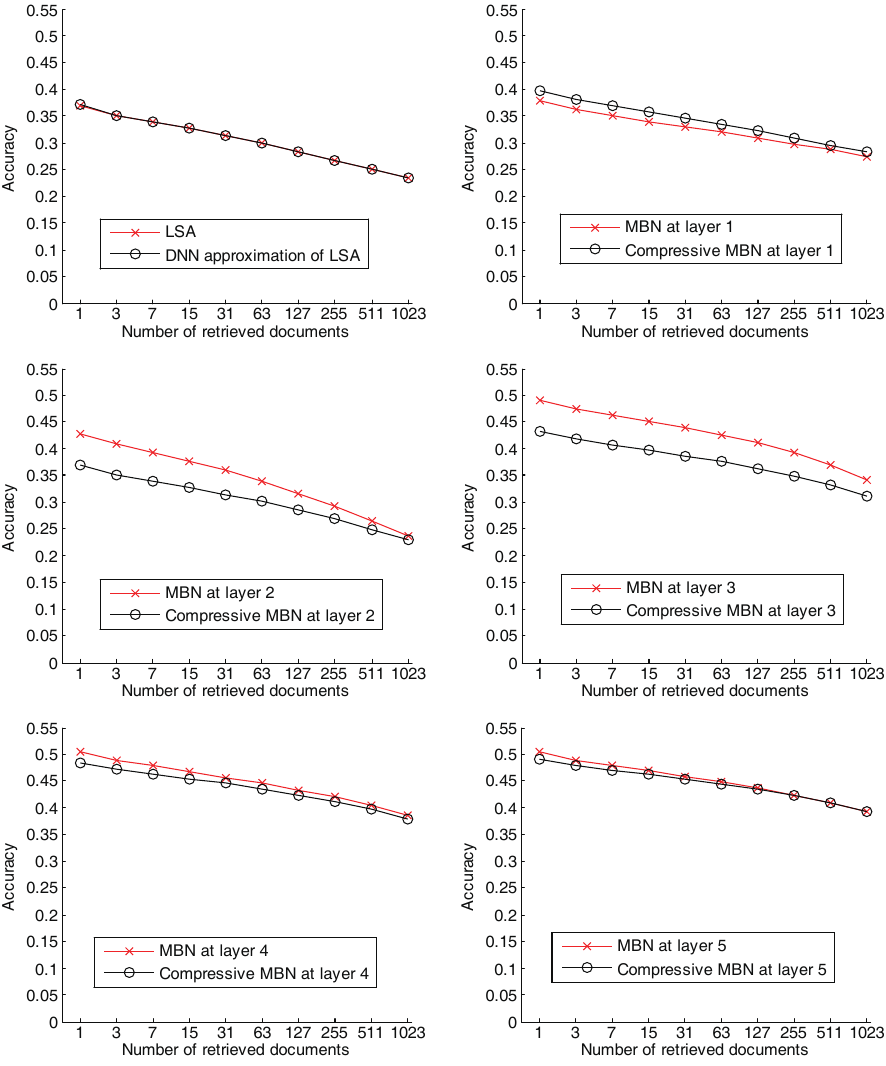}
 \caption{{Comparison of the generalization ability of MBN and the compressive MBN on retrieving the Reuters newsware stories.}}
  \label{fig:CMBN_RCV}
 \end{figure}

We then studied the generalization ability of the compressive MBN in retrieving the Reuters newsware stories. The neural network here has the same structure with that on MNIST. It projects the data space to the 5-dimensional representation produced by MBN.
The result in Fig. \ref{fig:CMBN_RCV} shows that the compressive MBN produces an almost identical accuracy curve with MBN. The prediction time of MBN and the compressive MBN is 190,574.74 and 88.80 seconds, respectively.

\section{Discussions}\label{sec:motivate}
MBN is related to many methods in statistics and machine learning. Here we introduce its connection to histogram-based density estimators, bootstrap methods, clustering ensemble, vector quantization, product of experts, sparse coding, and unsupervised deep learning.

\subsection{Histogram-based density estimators}
Histogram-based density estimation is a fundamental density estimation method, which estimates a probability function by accumulating the events that fall into the intervals of the function \citep{freedman2007statistics} where the intervals may have an equivalent length or not. Our proposed density estimator is essentially such an approach. An interval in our approach is defined as the data space between two data points. The accumulated events in the interval are the shared centroids by the two data points.

\subsection{Bootstrap methods}

Bootstrap resampling \citep{efron1979bootstrap,efron1993introduction} has been applied successfully to machine learning. It resamples data \textit{with replacement}. MBN does not adopt the standard bootstrap resampling. In fact, MBN uses random subsampling \textit{without replacement}, also known as \textit{delete-$d$ jackknife resampling} in statistics \citep{efron1993introduction}. The reason why we do not adopt the standard bootstrap resampling is that the resampling in MBN is used to build local coordinate systems, hence, if a data point is sampled multiple times, the duplicated data points are still viewed as a single coordinate axis. Moreover, it will cause the $k$-centroids clusterings built in different data spaces.
However, MBN was motivated from and shares many common properties with the bootstrap methods, such as building each base clustering from a random sample of the input and de-correlating the base clusterings by the random sampling of features \citep{breiman2001random} at each base clustering, hence, we adopted the phrase ``bootstrap'' in MBN and clarify its usage here for preventing confusion.

\subsection{Clustering ensemble}
Clustering ensemble \citep{strehl2003cluster,dudoit2003bagging,fern2003random,fred2005combining,zheng2010hierarchical,vega2011survey} is a clustering technique that uses a \textit{consensus function} to aggregate the clustering results of a set of mutually-independent base clusterings. Each base clustering is usually used to classify data to the ground-truth number of classes. An exceptional clustering ensemble method is \citep{fred2005combining}, in which each base $k$-means clustering produces a subclass partition by assigning the parameter $k$ to a random value that is slightly larger than the ground-truth number of classes.

Each layer of MBN can be regarded as a clustering ensemble. However, its purpose is to estimate the density of data instead of producing an aggregated clustering result. Moreover, MBN has clear theoretical and geometric explanations. We have also found that setting $k$ to a random value does not work for MBN, particularly when $k$ is also small.

\subsection{Vector quantization}
{\color{black}{Each layer of MBN can be regarded as a vector quantizer to the input data space. The codebook produced by MBN is exponentially smaller than that produced by a traditional $k$-means clustering, when they have the same level of quantization errors.
A similar idea, named \textit{product quantization}, has been explored in \citep{jegou2011product}. If product quantization uses random sampling of features instead of a fixed non-overlapping partition of features in \citep{jegou2011product}, and uses random sampling of data points to train each $k$-means clustering instead of the expectation-maximization optimization, then product quantization equals to a single layer of MBN. We are also aware of the hierarchical product quantizer \citep{wichert2012product}, which builds multiple sets of sub-quantizers (where the name ``hierarchy'' comes) on the original feature. Our MBN builds each layer of sub-quantizers on the output sparse feature of its lower layer which is fundamentally different from the hierarchical product quantization method.

 An important difference between existing vector quantization methods (e.g. binarized neural networks and binary hashing) and MBN is that the former are designed for reducing the computational and storage complexities, while MBN is not developed for this purpose. MBN aims at providing a simple way to overcome the difficulty of density estimation in a continuous space. Hence, it generates larger codebooks than common vector quantization methods and does not transform the sparse codes to compact binary codes.}}

  \subsection{Product of experts}\label{subsec:poe}
PoE aims to combine multiple individual models by multiplying them, where the individual models have to be a bit more complicated and each contains one or more hidden variables \citep{hinton2002training}. Its general probability framework is:
 \begin{eqnarray}\label{eq:poe}
p(\x)=\frac{\prod_{v=1}^{V}g_v(\x)}{\sum_{\x'}\prod_{v=1}^{V}g_v(\x')}
 \end{eqnarray}
 where $\x'$ indexes all possible vectors in the data space, and $g_{v}$ is called an \textit{expert}. A major merit of PoE is that, a function that can be fully expressed by a \textit{mixture of experts} with $N$ experts (i.e. mixtures), such as Gaussian mixture model or $k$-means clustering, can be expressed compactly by a PoE with only $\log_2 N$ experts at the minimum, with the expense of the optimization difficulty of the partition function ${\sum_{\x'}\prod_{v=1}^{V}g_v(\x')}$ which consists of an exponentially large number of components.

 The connection between MBN and PoE is as follows:
 \begin{thm}\label{theorem:3}
      Each layer of MBN is a PoE that does not need to optimize the partition function.
 \end{thm}
 \begin{proof}
 See \ref{app:b} for the proof.

 \end{proof}

\subsection{Sparse coding}
Given a learned dictionary $\W$, sparse coding typically aims to solve $\min_{\h_i} \sum_{i=1}^{n} \|\x_i-\W\h_i \|^2_2 + \lambda\| \h_i\|_1^1$,
where $\| \cdot\|_q$ represents $\ell_q$-norm, $\h_i$ is the sparse code of the data point $\x_i$, and $\lambda$ is a hyperparameter controlling the sparsity of $\h_i$. Each column of $\W$ is called a basis vector.

{\color{black}{To understand the connection between MBN and sparse coding}}, we may view $\lambda$ as a hyperparameter that controls the number of clusterings. Specifically, if we set $\lambda = 0$, it is likely that $\h_i$ contains only one nonzero element. Intuitively, we can understand it as that we use only one clustering to learn a sparse code. A good value of $\lambda$ can make a small part of the elements of $\h_i$ nonzero. This choice approximates to the method of partitioning the dictionary to several (probably overlapped) subsets and then grouping the basis vectors in each subset to a base clustering. Motivated by the above intuitive analysis, we introduce the connection between sparse coding and MBN formally as follows:
\begin{thm}\label{theorem:2}
The $\ell_{1}$-norm sparse coding is a convex relaxation of the building block of MBN when given the same dictionary.
\end{thm}
\begin{proof}
See \ref{app:d} for the proof.
\end{proof}

\subsection{Unsupervised deep learning}

Learning abstract representations by deep networks is a recent trend. From the geometric point of view, the abstract representations are produced by reducing larger and larger local variations of data from bottom up in the framework of trees that are built on data spaces.\footnote{Some recent unsupervised deep models for data augmentation are out of the discussion, hence we omit them here.} For example, convolutional neural network merges child nodes by pooling. Hierarchical Dirichlet process \citep{Teh05sharingclusters} builds trees whose father nodes generate child nodes according to a prior distribution. DBN \citep{hinton2006reducing} merges child nodes by reducing the number of the nonlinear units gradually from bottom up. Subspace tree \citep{wichert2015projection} merges nodes by reducing the dimensions of the subspaces gradually. PCANet \citep{chan2015pcanet} merges nodes by reducing the output dimensions of the local PCA associated with its patches gradually. Our MBN merges nodes by reducing the number of the randomly sampled centroids gradually.

A fundamental difference between the methods is how to build effective local coordinate systems in each layer. To our knowledge, MBN is a simple method and needs little assumption and prior knowledge. It is only more complicated than random projection which is, to our knowledge, not an effective method for unsupervised deep learning to date. Moreover, MBN is the only method working in discrete feature spaces, and it works well.

\section{Conclusions}\label{sec:5}
In this paper, we have proposed multilayer bootstrap network for nonlinear dimensionality reduction. MBN has a novel network structure that each expert is a $k$-centroids clustering whose centroids are randomly sampled data points with randomly sampled features; the network is gradually narrowed from bottom up.

MBN is composed of two novel components: (i) each layer of MBN is a nonparametric density estimator by random resampling. It estimates the density of data correctly without any model assumption. It is exponentially more powerful than a single $k$-centroids clustering. Its estimation error is proven to be small and controllable. (ii) The network is a deep ensemble model. It essentially reduces the nonlinear variations of data by building a vast number of hierarchical trees on the data space. It can be trained as many layers as needed with both large-scale and small-scale data.

  MBN performs robustly with a wide range of parameter settings. Its time and storage complexities scale linearly with the size of training data. It supports parallel computing naturally. Empirical results demonstrate its efficiency at the training stage and its effectiveness in density estimation, data visualization, clustering, and document retrieval. We have also demonstrated that the high computational complexity of MBN at the test stage can be eliminated by the compressive MBN---a framework of unsupervised model compression based on neural networks.

  A problem left is on the selection of parameter $\delta$ which controls the network structure. Although the performance of MBN with $\delta=0.5$ is good, there is still a large performance gap between $\delta=0.5$ and the best $\delta$. Hence, how to select $\delta$ automatically is an important problem.

\section*{Acknowledgement}

The author thanks Prof. DeLiang Wang for providing his computing resources. The author also thanks the action editor and reviewers for their comments which improved the quality of the paper.
This work was supported by the National Natural Science Foundation of China under Grant 61671381.

\section*{References}

\bibliography{zxlrefs,zxlrefs2} 

\appendix
  \section{Complexity analysis}\label{app:a1}

\begin{thm}\label{thm:3}
The computational and storage complexities of MBN are:
\setlength{\arraycolsep}{0.1em}
\begin{eqnarray}
  O_{time} &=& O\left(ds k Vn \right)\\
    O_{storage} &=& O((ds+V)n+kV+kds)
\end{eqnarray}
respectively at the bottom layer, and are:
\begin{eqnarray}
  O_{time} &=& O\left(k V^2n  \right)\\
    O_{storage} &=& O(2V(n+k))
\end{eqnarray}
respectively at other layers, where $d$ is the dimension of the original feature, $s$ is the sparsity of the data (i.e., the ratio of the non-zero elements over all elements).

If MBN is not used for prediction, then MBN needs not to be saved, which further reduces the storage complexity to $O((ds+V)n)$ at the bottom layer and $O( 2Vn)$ at other layers.
\end{thm}

\begin{proof}
Due to the length limitation of the paper, we omit the simple proof.
\end{proof}

\begin{figure*}[t]
 \centering
         \includegraphics[width=13cm]{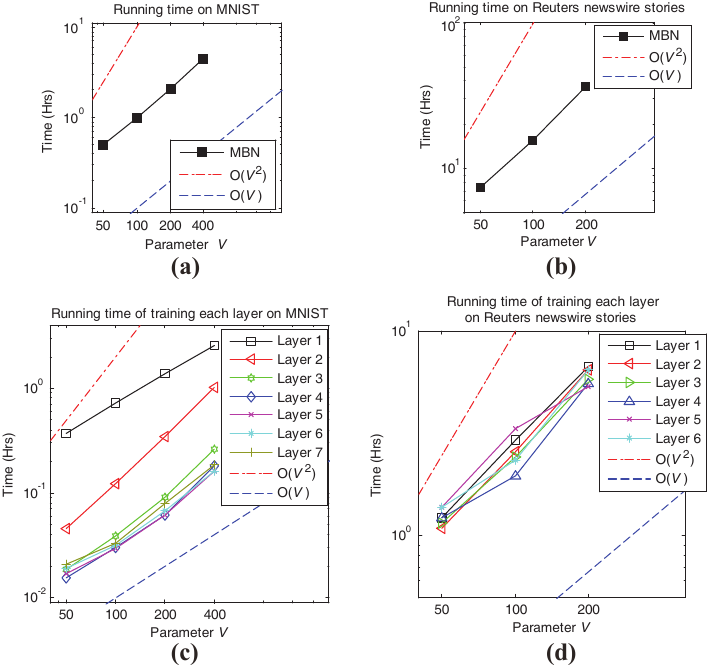}
         \caption{{Time complexity of MBN with respect to parameter $V$.}}
         \label{fig:supp5}
 \end{figure*}

Fortunately, as shown in Fig \ref{fig:supp5}, the empirical time complexity grows with $O(V)$ instead of $O(V^2)$. The only explanation is that the input data is sparse. Specifically, the multiplication of two sparse matrices only considers the element-wise multiplication of two elements that are both nonzero, as a result, when the input data is sparse, one factor $V$ is offset by the sparsity factor $s$. The empirical time and storage complexities with other parameters are consistent with our theoretical analysis. We omit the results here.

 Note that, our default parameter $k_1=0.5 n$ makes the time complexity scale squarely with the size of the data set $n$. To reduce the computational cost, we usually set $k_1$ manually to a small value irrelevant to $n$ as what we have done for visualizing the full MNSIT and retrieving the RCV1 documents in Section \ref{sec:4}.

\section{Proof of Theorem \ref{theorem:3}}\label{app:b}
The optimization objective of $k$-means clustering can be written as:
\setlength{\arraycolsep}{0.2em}
     \begin{eqnarray}
&&\max_{\W,\h}\mbox{ }g(\x|\W,\h)\nonumber\\
= &&\max_{\W,\h}\mbox{ }\exp\left(-\frac{1}{2\sigma^2}\|\x-\W^T\h\|^2_2\right)\nonumber\\
 \mbox{subject to}&&\h \mbox{ is a one-hot code}
 \end{eqnarray}
 where $\W = [\w_{1},\ldots,\w_{k}]$ is the weight matrix whose columns are centroids, $\h$ is a vector of hidden variables, and the covariance matrix of the clustering is $\sigma^2\mathbf{I}$ with $\mathbf{I}$ being the identity matrix and $\sigma \rightarrow 0$ \citep{bishop2006pattern}.

 Substituting $g(\x|\W,\h)=\exp\left(-\|\x-\W^T\h\|^2_2/2\sigma^2\right)$ to Eq. (\ref{eq:poe}) gets:
  \begin{eqnarray}
&&p(\x|\{W_v,\h_v\}_{v=1}^V) \nonumber\\ &=&\frac{\prod_{v=1}^V\exp\left(-\|\x-\W_v^T\h_v\|^2_2/2\sigma^2\right)}{\sum_{\x'}\prod_{v=1}^V\exp\left(-\|\x'-\W_v^T\h_v\|^2_2/2\sigma^2\right)}\nonumber\\
&=&\frac{\exp\left(-\sum_{v=1}^V\|\x-\W_v^T\h_v\|^2_2/2\sigma^2\right)}{\sum_{\x'}\exp\left(-\sum_{v=1}^V\|\x'-\W_v^T\h_v\|^2_2/2\sigma^2\right)}\label{eq:poe_kmeans}
 \end{eqnarray}
 where we assume that all experts have the same covariance matrix $\sigma^2\mathbf{I}$. Maximizing the likelihood of Eq. (\ref{eq:poe_kmeans}) is equivalent to the following problem:
   \begin{eqnarray}
\mathcal{J}_{\footnotesize\mbox{PoE}}
&=&\min_{\{W_v,\h_v\}_{v=1}^V}-\log p(\x|\{W_v,\h_v\}_{v=1}^V) \nonumber\\ &=&\min_{\{W_v,\h_v\}_{v=1}^V}{\frac{1}{2\sigma^2}\sum_{v=1}^V\|\x-\W_v^T\h_v\|^2_2}\nonumber\\
&&+\log{\sum_{\x'}\exp\left(-\frac{1}{2\sigma^2}\sum_{v=1}^V\|\x'-\W_v^T\h_v\|^2_2\right)}\nonumber\\
&\propto&\min_{\{W_v,\h_v\}_{v=1}^V}{\sum_{v=1}^V\|\x-\W_v^T\h_v\|^2_2}\nonumber\\
&&+\log\left({\sum_{\x'}\exp\left(-\frac{1}{2\sigma^2}\sum_{v=1}^V\|\x'-\W_v^T\h_v\|^2_2\right)}\right)^{2\sigma^2}\nonumber\\
&& \mbox{subject to}\quad \h_v \mbox{ is a one-hot code}
\label{eq:poe_approx}
 \end{eqnarray}
 Because $\sigma \rightarrow 0$, problem (\ref{eq:poe_approx}) can be rewritten as:
    \begin{eqnarray}
\mathcal{J}_{\footnotesize\mbox{PoE}}
&\propto&\min_{\{W_v,\h_v\}_{v=1}^V}{\sum_{v=1}^V\|\x-\W_v^T\h_v\|^2_2}\nonumber\\
&&+\log\left({\max_{\x'}\left(\exp\left(-\frac{1}{2\sigma^2}\sum_{v=1}^V\|\x'-\W_v^T\h_v\|^2_2\right)\right)}\right)^{2\sigma^2}\nonumber\\
&=&\min_{\{W_v,\h_v\}_{v=1}^V}{\sum_{v=1}^V\|\x-\W_v^T\h_v\|^2_2}-\min_{\x'}\sum_{v=1}^V\|\x'\nonumber\\
&&-\W_v^T\h_v\|^2_2\nonumber\\
&&\mbox{subject to} \quad \h_v \mbox{ is a one-hot code}
\label{eq:poe_min}
 \end{eqnarray}
 Because $\x'$ can be any possible vector in the data space, we have $\min_{\x'}\sum_{v=1}^V\|\x'-\W_v^T\h_v\|^2_2 = 0$. Eventually, the optimization objective of PoE is:
     \begin{eqnarray}
\mathcal{J}_{\footnotesize\mbox{PoE}}
&\propto&\min_{\{W_v,\h_v\}_{v=1}^V}{\sum_{v=1}^V\|\x-\W_v^T\h_v\|^2_2}\nonumber\\
\mbox{subject to}&&\h_v \mbox{ is a one-hot code}
\label{eq:poe_final}
 \end{eqnarray}
 which is irrelevant to the partition function.

It is clear that problem (\ref{eq:poe_final}) is the optimization objective of an ensemble of $k$-means clusterings. When we assign $W_v$ by random sampling, then  problem (\ref{eq:poe_final}) becomes:
\begin{eqnarray}
\mathcal{J}_{\footnotesize\mbox{PoE}}
&\propto&\min_{\{\h_v\}_{v=1}^V}{\sum_{v=1}^V\|\x-\W_v^T\h_v\|^2_2}\nonumber\\
\mbox{subject to}&&\h_v \mbox{ is a one-hot code}
\label{eq:poe_final2}
 \end{eqnarray}
 which is the building block of MBN. Theorem \ref{theorem:3} is proved.

\section{Proof of Theorem \ref{theorem:2}}\label{app:d}

\setlength{\arraycolsep}{0.1em}
Each layer of MBN maximizes the likelihood of the following equation:
  \begin{eqnarray}
p(\x)={\prod_{v=1}^{V}g_v(\x)}
 \label{eq:rbm}
 \end{eqnarray}
where $g_v(\x)$ is a $k$-means clustering with the squared error as the similarity metric:
    \begin{eqnarray}
g_v(\x) &= &\mathcal{MN}\left( \x; \W_v\h_v,\sigma^2\mathbf{I}\right) \label{eq:k-means}\\
 \mbox{subject to} &&\h_v \mbox{ is a one-hot code} \nonumber
 \end{eqnarray}
 where $\mathcal{MN}$ denotes the multivariate normal distribution, $\W_v = [\w_{v,1},\ldots,\w_{v,k}]$ is the weight matrix whose columns are centroids, $\mathbf{I}$ is the identity matrix, and $\sigma \rightarrow 0$.

  Given a data set $\{\x_i\}_{i=1}^n$. We assume that $\{\W_v\}_{v=1}^V$ are fixed, and take the negative logarithm of Eq. ({\ref{eq:rbm}}):
  \begin{eqnarray}
\min_{\{\{\h_{v,i}\}_{i=1}^{n}\}_{v=1}^V}&\mbox{ }& \sum_{v=1}^{V}\sum_{i=1}^{n} \|\x_i-\W_v\h_{v,i} \|^2_2, \label{eq:object}\\
 \mbox{subject to} &\mbox{ }&\h_{v,i} \mbox{ is a one-hot code}.\nonumber
 \end{eqnarray}
 If we denote $\W = [\W_1,\ldots,\W_V]$ and further complement the head and tail of $\h_{v,i}$ with multiple zeros, denoted as $\h_{v,i}'$, such that $\W\h_{v,i}' = \W_v\h_{v,i}$, we can rewrite Eq. ({\ref{eq:object}}) to the following equivalent problem:
  \begin{eqnarray}
\min_{\{\{\h_{v,i}\}_{i=1}^{n}\}_{v=1}^V}&\mbox{ }& \sum_{v=1}^{V}\sum_{i=1}^{n} \|\x_i-\W\h'_{v,i} \|^2_2, \label{eq:object2}\\
 \mbox{subject to} &\mbox{ }&\h_{v,i} \mbox{ is a one-hot code}.\nonumber
 \end{eqnarray}
 It is an integer optimization problem that has an integer matrix variable $\H'_v =[\h'_{v,1},\ldots,\h'_{v,n}]$. Suppose there are totally $|\mathcal{H}'_{v}|$ possible solutions of $\H'_v$, denoted as $\H'_{v,1},\ldots,\H'_{v,|\mathcal{H}'_{v}|}$, we first relax Eq. ({\ref{eq:object2}}) to a convex optimization problem by constructing a \textit{convex hull} \citep{boyd2004convex} on $\H'_v$:
 \begin{eqnarray}
\min_{\left\{ \{\mu_{v,k}\}_{k=1}^{|\mathcal{H}'_v|}  \right\}_{v=1}^V} &\mbox{ }&  \sum_{v=1}^{V}\sum_{i=1}^{n} \left\|\x_i-\W\left(\sum_{k=1}^{|\mathcal{H}'_v|}\mu_{v,k}\h'_{v,k,i}\right) \right\|^2_2  \label{eq:object2_hull} \\
\mbox{subject to}&\mbox{ }& 0\le\mu_{v,k}\le 1,\sum_{k=1}^{|\mathcal{H}'_v|}\mu_{v,k}=1,\quad\forall v = 1,\ldots,V.\nonumber
 \end{eqnarray}
Because Eq. ({\ref{eq:object2_hull}}) is a convex optimization problem, according to Jensen's inequality, the following problem learns a lower bound of Eq. ({\ref{eq:object2_hull}}):
 \begin{eqnarray}
\min_{\left\{ \{\mu_{v,k}\}_{k=1}^{|\mathcal{H}'_v|}  \right\}_{v=1}^V} &\mbox{ }&  V\sum_{i=1}^{n} \left\|\x_i-\W\h_i'' \right\|^2_2 \label{eq:object3} \\
\mbox{subject to} &\mbox{ }& 0\le\mu_{v,k}\le 1,\sum_{k=1}^{|\mathcal{H}'_v|}\mu_{v,k}=1,\quad \forall v = 1,\ldots,V\nonumber
 \end{eqnarray}
 where $\h_i'' = \frac{1}{V}\sum_{v=1}^V\sum_{k=1}^{|\mathcal{H}'_v|}\mu_{v,k}\h_{v,k,i}'$ with $\mu_{v,k}$ as a variable.

  Recalling the definition of sparse coding given a fixed dictionary $\W$,
 we observe that Eq. ({\ref{eq:object3}}) is a special form of sparse coding with more strict constraints on the format of sparsity.

Therefore, given the same dictionary $\W$, each layer of MBN is a distributed sparse coding that is lower bounded by the common $\ell_1$-norm sparse coding.
When we discard the expectation-maximization optimization of each $k$-means clustering (i.e., dictionary learning) but only preserve the default initialization method -- random sampling, Eq. ({\ref{eq:rbm}}) becomes the building block of MBN. Given the same dictionary, the $\ell_1$-norm-regularized sparse coding is a convex relaxation of the building block of MBN. Theorem \ref{theorem:2} is proved.

\section{Visualizations produced by intermediate layers}\label{app:c}



\clearpage
\begin{figure*}
         \center\includegraphics[width=12cm]{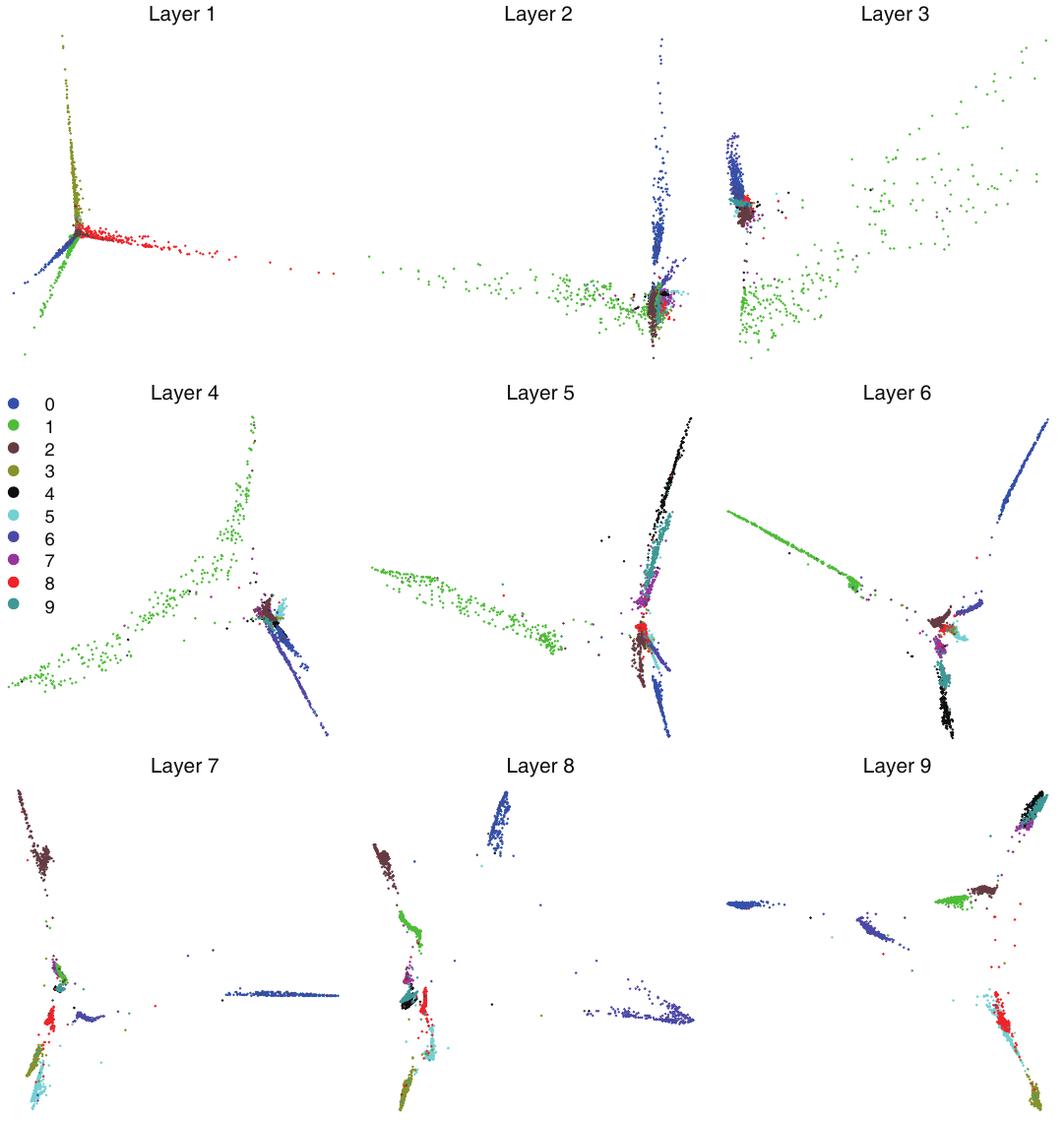}
         \caption{{Visualizations of MNIST produced by MBN at different layers. This figure is a supplement to Fig. \ref{fig:MNIST_visualization}.}}
         \label{fig:supp2}
 \end{figure*}

\end{document}